\documentclass[]{spie}

\usepackage[cmex10]{amsmath}
\usepackage{amssymb}
\usepackage{color}

\usepackage{wasysym}

\newtheorem{prop}{Proposition}[section]

\def\tT{{\mbox{\tiny{T}}}}
\newcommand{\bitem}{\begin{itemize}}
\newcommand{\eitem}{\end{itemize}}

\newcommand{\bpm}{\begin{pmatrix}}   
\newcommand{\epm}{\end{pmatrix}}

\newcommand{\bq}{\begin{equation}}
\newcommand{\eq}{\end{equation}}

\newcommand{\R}{\mathcal{R}}

\let\abs=\envert

\let\norm=\enVert

\let\inprod=\inProd

\DeclareMathOperator{\diver}{div}

\usepackage{graphicx}
\usepackage{bm}
\usepackage{algorithmic}

\usepackage{color}

\usepackage[tight,footnotesize]{subfigure}

\usepackage{epsfig}
\usepackage{epstopdf}
\begin{document}
%
 
\title{RANCOR: Non-Linear Image Registration with Total Variation Regularization}

%
%
%
\author{Martin Rajchl, John~S.H.~Baxter, Wu~Qiu, Ali~R.~Khan, 
Aaron~Fenster, Terry~M.~Peters, and~Jing~Yuan 
\thanks{All authors are with the Imaging Laboratories, Robarts Research Institute, Western University,
London, ON, N6A5K8, Canada}
\thanks{Corresponding author: Martin Rajchl e-mail: mrajchl@robarts.ca}
}

\maketitle

\begin{abstract}
Optimization techniques have been widely used in deformable registration, allowing for the incorporation of similarity metrics
with regularization mechanisms. These regularization mechanisms are designed to mitigate the effects of trivial solutions
to ill-posed registration problems and to otherwise ensure the resulting deformation fields are well-behaved. This 
paper introduces a novel deformable registration algorithm, RANCOR, which uses iterative convexification to address
deformable registration problems under total-variation regularization. Initial comparative results against four
state-of-the-art registration algorithms are presented using the Internet Brain Segmentation Repository (IBSR) database.

\end{abstract}

\begin{keywords}
Deformable Image Registration, Total Variation Regularization, Brain, Convex Relaxation, GPGPU
\end{keywords}

%

%
%
%
%

 \section{Introduction}
Registration is the systematic spatial deformation of one medical image as to allign it with another, either in the
same or a different modality. Although registration of images common to a single patient can rely largely on rigid
transformations, registration between patient images, common in techniques such as atlas construction or atlas-based
segmentation, have relied on highly non-linear (NLR) deformations often in the absence of highly detectable and
localizable landmarks. Non-linear registration aims to address these problems, using a similiarity metric to
judge the quality of the alignment after deformation, and a regularization mechanism to ensure that the deformation
field avoid trivial and otherwise undesirable components such as gaps or singularities. Non-linear registration is
indeed a challenging problem with many competing facets. Our algorithm is intended to provide an additional
option that could facilitate atlas building and segmentation techniques.

Many components of our deformable registration algorithm display a large amount of inherent parallelism between image
voxels. Such algorithms have been of growing interest to the medical imaging community because of the ability to
implement them on commercially avialable general purpose graphics processing units (GPGPUs) to dramatically improve
their speed and computational efficiency for both registration\cite{modat2010fast} and segmentation problems
\cite{rajchl2012fast, qiu2013fast, rajchl2014interactive}.

\subsection{Contributions}
We propose a novel non-linear RegistrAtioN via COnvex Relaxation (RANCOR) algorithm that allows for the combination of any pointwise error metric (such as the sum of absolute intensity differences (SAD) for intra-modality registration, and mutual information (MI) \cite{wells1996multi,rogelj2003point} or modality independent neighbourhood descriptors (MIND) \cite{heinrich2012mind} for inter-modality registration) while regularizing the deformation field by its total variation. This algorithm is implemented on GPGPU to ensure high performance.

\subsection{Previous studies}
Recent surveys provide a good overview of existing NLR registration methods \cite{crum2004non,gholipour2007brain} and we would like to emphasize the study performed by Klein et al. \cite{klein2009evaluation}, where 14 NLR registration algorithms were compared across four open brain image databases. We will compare our proposed method against the four highest ranked methods identified in \cite{klein2009evaluation}:

\noindent {\bf \emph{Advanced Normalization Tools (ANTs):}}
The \emph{Symmetric Normalization (SyN)} NLR method in \cite{avants2008symmetric} uses a multi-resolution
scheme to enforce a bi-directional diffeomorphism while maximizing a cross-correlation metric. It has been
shown in several open challenges \cite{farahani2013multimodal,klein2009evaluation,murphy2011evaluation}
to outperform well established methods. \emph{SyN} regularizes the deformation field through Gaussian
smoothing and enforcing transformation symmetry.

\noindent {\bf \emph{Image Registration Roolkit (IRTK):}}
The well-known \emph{Fast Free-Form deformations (F3D)} method in \cite{rueckert1999nonrigid} defines a
lattice of equally spaced control points over the target image and, by moving each point, locally modifies
the deformation field. Normalized mutual information combined with a cubic b-spline bending energy is
used as the objective function. Its multi-resolution implementation employs coarsely-to-finely spaced
lattices and Gaussian smoothing.

\noindent {\bf \emph{Automatic Registration Toolbox (ART):}}
\cite{ardekani2005quantitative} presents a homeomorphic NLR method using normalized cross-correlation
as similarity metric in a multi-resolution framework. The deformation field is regularized via median
and low-pass Gaussian filtering.

\noindent {\bf \emph{Statistical Parametric Mapping DARTEL Toolbox (SPM\_D):}}
The \emph{DARTEL} algorithm presented in \cite{ashburner2007fast} employs a static finite difference model
of a velocity field. The flow field is considered as a member of the Lie algebra, which is exponentiated
to produce a deformation inherently enforcing a diffeomorphism. It is implemented in a recursive,
multi-resolution manner.
 \section{Methods}
          
In this section, we propose a multi-scale dual optimization based method to
estimate the non-linear deformation field $u(x) = [u_1(x), u_2(x), u_3(x)]^\tT$,
bewteen two given images $I_1(x)$ and $I_2(x)$, which explores the minimization of the variational optical-flow energy
function:
\bq \label{eq:eng-func}
\min_u \;\; P(I_1, I_2; u) \, + \, R(u)
\eq
where the function term $P(I_1, I_2; u)$ stands for a dissimilarity measure of
the two input images $I_1(x)$ and $I_2(x)$ under deformation by $u$, and $R(u)$ gives the regularization function to
single out a smooth deformation field.
In this paper, we use the sum of absolute intensity differences (SAD):
\bq \label{eq:01}                 
P(I_1, I_2; u) \, := \, \int_{\Omega} \abs{I_1(x+u) - I_2(x)}\, dx 
\, ,
\eq  
as a simple similarity metric for two input images from the same modality.
%
The proposed framework can also be directly adapted for more advanced
image dissimilarity measures designed for registration between different modalities.
%

A regularization term, $R(u)$,  is
often incorporated to make the minimization problem \eqref{eq:eng-func} well-posed.
Otherwise, minimizing the image dissimilarity function $P(I_1, I_2; u)$ can result in 
trivial or infinite solutions. 
We consider the total variation of the deformation field as the regularization term:
\bq \label{eq:lreg}
R(u) \, := \alpha \int_\Omega \left( \abs{\nabla u_1}+\abs{\nabla u_2}+\abs{\nabla u_3} \right) dx \text{ .}
\eq  

   
The expected non-convexity of $I_1(x)$ and $I_2(x)$,
 makes it challenging to directly minimize \eqref{eq:eng-func}, even with
convex regularization. To address this issue, we introduce an incremental 
convexification approach, which lends itself to a standard coarse-to-fine
framework and allows for a more global perspective and
avoiding local optima by capturing large deformations.

In Section \ref{sec:c2f}, we develop the multi-scale optimization framework,
developing a sequence of related minimization problems. Each of these problems
are solved through a new non-smooth Gauss-Newton (GN) approach introduced in Section \ref{sec:scdo}.
which employs a novel sequential convexification and dual optimization procedure. 

\subsection{Coarse-to-Fine Optimization Framework} \label{sec:c2f}

The first stage in our approach is the construction of the image pyramid.
Let $I_1^1(x)$ \ldots $I_1^L(x)$ be
the $L$-level pyramid representation of $I_1(x)$ from
the coarsest resolution $I_1^1(x)$ to the finest resolution $I_1^L(x)=I_1(x)$,
and $I_2^1(x)$ \ldots $I_2^L(x)$ the $L$-level coarse-to-fine pyramid
representation of $I_2(x)$.
The optimization process is started from the coarsest level,
$\ell=1$, which extracts the deformation field $u^{1}(x)$ between 
$I_1^1(x)$ and $I_2^1(x)$ such that:
\bq \label{eq:1-level}
\min_{u^{1}} \;\; P(I_1^{1}(x), I_2^{1}(x); u^{1}) \, +
\, R(u^{1}) \, .
\eq
The vector field $u^1(x)$ gives the optimal deformation field
at the coarsest scale. It is warped to the next finer-resolved level,
$\ell=2$, to compute the optimal finer-level defomration field $u^2(x)$.
The process is repeated, obtaining the deformation field
$u^3(x)$ \ldots $u^L(x)$ at each level sequentially.

Second, at each resolution level $\ell$,
$\ell=2 \ldots L$, we compute an incremental deformation field $t^{\ell}(x)$
based on the two image functions $I_2^{\ell}(x)$ and $I_1^{\ell}(x+u^{\ell-1})$, 
where $I_1^{\ell}(x+u^{\ell-1})$ is warped by the deformation field
$u^{\ell-1}(x)$ computed at the previous resolution level $\ell-1$, i.e.
\bq \label{eq:each-level}
\min_{t^{\ell}} \;\; P(I_1^{\ell}(x+u^{\ell-1}), I_2^{\ell}(x); t^{\ell}) \, +
\, R(u^{\ell-1} + t^{\ell}) \, .
\eq
Clearly, the optimization problem \eqref{eq:1-level} can be viewed as the
special case of \eqref{eq:each-level}, i.e. for $\ell=1$, we define $u^0(x) = 0$
and $u^1(x) = (u^0 + t^1)(x)$. Therefore, the proposed coarse-to-fine
optimization framework sequentially explores the minimization of
\eqref{eq:each-level} at each image resolution level, from the coarsest $\ell=1$
to the finest $\ell=L$.
 
 \subsection{Sequential Convexification and Dual Optimization} \label{sec:scdo}
 
Now we consider the optimization problem \eqref{eq:each-level} for each image
resolution level $\ell$.
Given the highly non-linear function $P(I_1^{\ell}(x+u^{\ell-1}), I_2^{\ell}(x);
t^{\ell})$ in \eqref{eq:each-level}, we introduce a sequential
linearization and convexification procedure for this challenging
non-linear optimization problem \eqref{eq:each-level}. This results in 
a series of incremental warping steps in which each
step approximates an update of the deformation field $t^{\ell}(x)=
(t_1^{\ell}(x), t_2^{\ell}(x), t_3^{\ell}(x))^\tT$, until the updated deformation
is sufficiently small, i.e., it iterates through the following sequence of convex minimization steps
until convergence is attained:
\bitem
\item Initialize $(h^{\ell})^0(x) = 0$ and let $k=1$;
\item At the $k^{\text{th}}$ iteration, define the deformation field as
\[
\tilde{u}^{\ell-1}(x) \, := \, \Big(u^{\ell-1} + \sum_{i=0}^{k-1}
(h^{\ell})^i \Big)(x)
\] 
and compute the update deformation $(h^{\ell})^k$ to $\tilde{u}^{\ell-1}(x)$ by
minimizing the following convex energy function:
\bq \label{eq:la}
\min_{(h^{\ell})^k} \; \int_{\Omega} \abs{\tilde{P}^k_0 + \nabla \tilde{P}^k
\cdot (h^{\ell})^k} dx \, + \, R(u^{\ell-1} + (h^{\ell})^k) \, ,
\eq
where
\[
\tilde{P}^k((h^{\ell})^k) \, = \,  P(I_1^{\ell}(x+\tilde{u}^{\ell-1}),
I_2^{\ell}(x); (h^{\ell})^k) 
\]
and $\tilde{P}_0^k(x) = P(I_1^{\ell}(x+\tilde{u}^{\ell-1}),
I_2^{\ell}(x); 0) $. 
\item Let $k=k+1$ and repeat the second step till the new update $(h^{\ell})^k$
is small enough. Then, we have the total incremental deformation field
$t^{\ell(x)}$ at the image resolution level $\ell$ as:
\[
t^{\ell}(x) \, =\, \sum_{i=0}^{k} (h^{\ell})^i(x) \, .
\]
\eitem

These steps can be viewed as a non-smooth GN method for the non-linear optimization problem
\eqref{eq:each-level}, in contrast to the classical GN method
proposed in \cite{baust2010diffusion}. 
Moreover, the $L_1$-norm and the convex
regularization term $\R(\cdot)$, \eqref{eq:la} results in a convex optimization
problem. The non-smooth $L_1$-norm from \eqref{eq:la}
provides more robustness in practice than the conventional smooth $L_2$-norm
used in the classical GN method.

Solving the convex minimization problem \eqref{eq:la} is the most essential
step in the proposed algorithmic framework The introduced
primal-dual variational analysis not only provides an equivalent dual formulation to the
optimization problem \eqref{eq:la} but also derives an efficient
solution algorithm. First, we simplify the expression of the convex problem \eqref{eq:la} as:
\bq \label{eq:las}
\min_{h} \; \int_{\Omega} \abs{P_0 + \nabla P
\cdot h} dx \, + \, R(\tilde{u} + h) \, ,
\eq
where $\tilde{u}(x)$ represents the deformation field.
Through variational analysis, we
can derive an equivalent 
\emph{dual model} to \eqref{eq:las}:
\vspace{0mm}
\begin{prop} \label{prop:01}
The \emph{convex minimization problem} \eqref{eq:las} can be represented
by its \emph{primal-dual model} \eqref{eq:primal-dual} and \emph{dual model}:
\begin{align} \label{eq:dual}
\hspace{-0.4cm}\max_{\abs{w(x)}\leq 1, q} \, E(w,q) :=  & \int (w P_0  + \sum_{i=1}^3 \tilde{u}_i \diver
q_i) dx \, - \, R^{\ast}(q)
\end{align}
subject to     
\bq \label{eq:dual-const}
F_i(x)  :=  (w \cdot \partial_i P  +  \diver q_i)(x)  =  0 \, , \quad i = 1,2,3
\, .
\eq   
The dual regularization function $R^{\ast}(q)$ is given by \eqref{eq:dual-r1}.
\end{prop}

The proof is given in Appendix \ref{sec:proof01}.
 \vspace{3mm}

As shown in Appendix \ref{sec:proof01}, each component of the deformation field
$[h_1(x), h_2(x), h_3(x)]^\tT$ works as the optimal multiplier functions to their
respective constraints, \eqref{eq:dual-const}.
Therfore, the energy function of the primal-dual model \eqref{eq:primal-dual} 
is exactly the Lagrangian function to the
\emph{dual model} \eqref{eq:dual}:
\begin{align}
\nonumber L(h,w,q) = & E(w,q) \, + \, \sum_{i=1}^3 \inprod{h_i, F_i} \, ,
\end{align}      
where $E(w,q)$ and the linear functions $F_i(x)$, $i=1,2,3$, are defined in
\eqref{eq:dual} and \eqref{eq:dual-const} respectively. We can now 
derive an efficient duality-based Lagrangian augmented
algorithm based on the modern convex optimization theories (see
\cite{bertsekas1999nonlinear,yuan2010study,yuan2010continuous} for details), using
the augmented Lagrangian function:
\begin{align}     
L_c(h,w,q) \, = \, L(h,w,q) - \frac{c}{2} \sum_{i=1}^3 \norm{F_i}^2 \, ,
\end{align}         
where $c>0$ is a positive constant and the additional quadratic penalty function
is applied to ensure the functions \eqref{eq:dual-const} vanish. 
Our proposed \emph{duality-based optimization algorithm} is:
\bitem
\item Set the initial values of $w^0$, $q^0$ and $h^0$, and let $k=0$.
\item Fix $q^k$ and $h^k$, optimize $w^{k+1}$ by
\begin{align}
w^{k+1} \, :=\, \arg \max_{\abs{w(x)}\leq 1} \; L_c(h^k, w, q^k)  \,
\end{align}   
generating the convex minimization problem:
\begin{align}
\min_{\abs{w(x)}\leq 1} \int w P_0 dx + \frac{c}{2}
\sum_{i=1}^3 \int (w \partial_i P - T_i^k)^2 dx\, ;
\end{align}   
where $T_i^k(x)$ ($i=1,2,3$) is computed from the fixed variables $q^k$ and
$h^k$.
$w^{k+1}$ is computed by threshholding:
\begin{align}   
w^{k+1} \, = \, \mathrm{Threshhold}_{\abs{w(x)} \leq 1}(w^{k+1/2}(x))\, ,
\end{align}   
where
\[
w^{k+1/2} \, = \, \frac{c \sum_{i=1}^3 (\partial_i P \, \cdot
T_i^k) - P_0}{ c \sum_{i=1}^3 (\partial_i P)^2} \, .
\]
\item Fixing $w^{k+1}$ and $h^k$, optimize $q^{k+1}$ by
\begin{align}
q^{k+1} \, := \, \arg\min_{q} \; L_c(h^k, w^{k+1}, q) \, ;
\end{align} 
which amounts to three convex minimization problems:
\begin{align} \nonumber
& \min_{q_i} \, \int q_i \cdot \nabla \tilde{u}_i dx + \frac{c}{2}
\int (\diver q_i - U_i^k)^2 dx + R^{\ast}(q)\, ;\\
& i=1,2,3  \, ;
\end{align}   
where $U_i^k$ is computed from the fixed
variables $w^{k+1}$ and $h^k$.
Hence, $q_i^{k+1}$,
$i=1,2,3$, can be approximated by a
gradient-projection step corresponding to \eqref{eq:dual-r1}.
\item Once $w^{k+1}$ and $q^{k+1}$ are obtained, update $h^{k+1}$ by     
\begin{align}   \nonumber
&h_i^{k+1} \, = \, h^k - c\Big(w^{k+1} \cdot \partial_i P  +  \diver q_i^{k+1} \Big) \, ;\\
& i=1,2,3  \, ;
\end{align}         
\item Increment $k$ and iterate until converged, i.e.
\begin{align}              
c\int \abs{w^{k+1} \cdot \partial_i P  +  \diver q_i^{k+1}} dx  \le \delta \, ,        
\end{align}                                          
where $\delta$ is a chosen small positive parameter ($5\times10^{-4}$).
\eitem                  
                              
 \section{Experiments}

\subsection{Image Database}
The image data constisted of an open multi-center T1w MRI dataset with corresponding manual segmentations,
the Internet Brain Segmentation Repository (IBSR) database, totalling 18 labeled image volumes at 1.5T
available on \texttt{www.mindboggle.info} in a pre-processed form with labeling protocols and
transforms into MNI space.

The experiments were performed in a pair-wise manner. For each image in the database, seventeen registrations were performed using the chosen image as the reference image and one of the remaining as the floating image. Thus, our experiment consisted of 306 registration problems in total.
	
\subsection{Initialization \& Pre-processing}
Prior to registration, all images were skull stripped by constructing brain masks from manual labels
using morphological operations \cite{klein2009evaluation} and then affinely registered using the
\emph{FMRIB Software Library's (FSL)} \emph{FLIRT} package \cite{jenkinson2002improved} into the
space of the \emph{MNI152\_T1\_1mm\_brain}. These affine transformations were made available
on \texttt{www.mindboggle.info} and used to initialize the NLR algorithms. This guarantees that
the same initialization is used for the algorithms in \cite{klein2009evaluation} and allows
for quantitative comparisons.
As a pre-processing step, both affinely registered images were robustly normalized to zero mean
and standard deviation units to ensure a constant regularization weight $\alpha$ could be used.

\subsection{Implementation \& Parameter Tuning}
The proposed NLR method was implemented in MATLAB (Natick, MA) using the Compute Unified Device
Architecture (CUDA) (NVIDIA, Santa Clara, CA) for GPGPU computing. Each level in the coarse-to-fine
framework constists of multiple warps invoking the proposed GPGPU accelerated regularization algorithm.
Parameter tuning of the regularization weight $\alpha$ was done on two randomly picked dataset pairs
similar to the tuning in \cite{klein2009evaluation}. All other parameters, such as the number of
levels ($N_{Levels}$), the number of warps ($N_{Warps}$) and the maximum number of iterations
($It_{MAX}$) were determined heuristically on a single image volume not used in this study.
Table~\ref{ta:parameters} contains all set parameter values.

\begin{table}[th!]\caption{\label{ta:parameters} Registration algorithm parameters}
\label{ta:parameters}
\centering
\begin{tabular}{lcccc}
\hline
\textbf{Method} & \textbf{$\alpha$} & \textbf{$N_{Levels}$} & \textbf{$N_{Warps}$} & \textbf{$It_{MAX}$} \\
\hline
RANCOR TVR & 0.30 & 3 & 4 & 220 \\
\hline
\multicolumn{5}{l}{All parameters were kept constant across all experiments.}\\
\hline\hline
\end{tabular}
\end{table}

\subsection{Evaluation Metric}
To compare our registration method against other NLR registration algorithms, we used the
\emph{target overlap} (\emph{TO}) as a regional metric:
\begin{equation} \label{eq:target_overlap}
TO = \frac{\sum_L{|F_L \cup R_L|}}{\sum_L{|R_L|}}
\end{equation}
where $F$ is the floating image, $R$ the reference image, and $L$ a labeled region, as
indicated in \cite{klein2009evaluation}. This parallels our motivation of using NLR
registration to port segmentation labels
to incoming datasets, and takes advantage of the manual segmentations providing in the
IBSR database.

Results were considered significant if the probability of
making a type I error was less than 1\% ($p < 0.01$). For this purpose, we employed a series of two-tailed,
pairwise Student's t-test.
 \section{Results}

\subsection{Run times}
The experiments were conducted on a Ubuntu 12.04 (64-bit) desktop machine with 144 GB memory and an
NVIDIA Tesla C2060 (6 GB memory) graphics card. The maximum run times for the MATLAB code
including pre-processing, optimization, and GPGPU enhanced regularization are given in
Table~\ref{ta:runtimes}.

\begin{table}[ht!]\caption{\label{ta:runtimes}Maximum GPGPU TV regularization run times at each level $l$ and
total registration time in seconds. }
\label{ta:Acc_TO}
\centering
\begin{tabular}{l|ccc|c}
\hline
\hline
&   \textbf{TVR $(l=0)$}  & \textbf{TVR $(l=1)$} & \textbf{TVR $(l=2)$} & \textbf{Total run time} \\
\hline
RANCOR TVR  & 0.25  &  1.76 & 13.58 & 76.76  \\ [1pt]
\hline \hline
\end{tabular}
\end{table}

\subsection{Accuracy}
\begin{figure}
\centering
\includegraphics[width=0.45\linewidth]{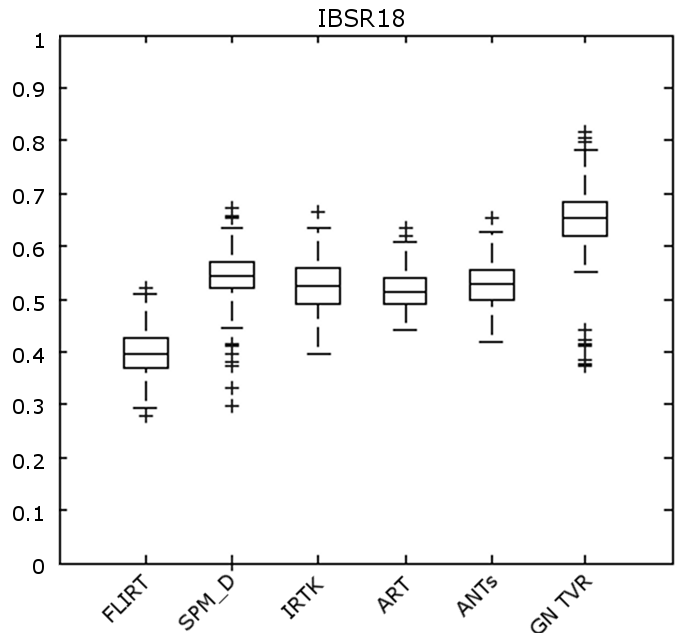}
\vspace{-2mm}

\caption{\label{fig:TO_all} Mean Target Overlap Results}
\end{figure}

Figure~\ref{fig:TO_all} shows boxplots of the TO accuracy for each of the registration methods. The
results were averaged across all regions. Numerical results are provided in Table~\ref{ta:Acc_TO}.

\begin{table}[ht!]\caption{\label{ta:Acc_TO}Mean target overlap (TO) accuracy}
\centering
\begin{tabular}{lcccc}
\hline
\hline
&  \textbf{IBSR18} \\
\hline
FLIRT    & $39.7\pm4.1$  \\
SPM\_D   & $54.0\pm4.7$ \\
IRTK     & $52.1\pm3.5$ \\
ART      & $51.5\pm3.5$\\
Syn      & $52.8\pm4.2$\\
RANCOR TVR & $64.9\pm6.3$\\ 
\hline \hline

\end{tabular}
\end{table}

 \section{Discussion}
We proposed a novel GPGPU-accelerated deformation field regularization method, total variation-based
regularization, for NLR registration. This method was implemented within a coarse-to-fine optimization
framework and compared on an open and publicly available database, IBSR. We employed the same initialization,
tuning conditions, and evaluation scripts to quantitatively compare the proposed methods against
four well-known NLR methods. Further, we numerically state the accuracy metrics allowing for direct
comparison.

The proposed method significantly outperformed the comparative methods in terms of TO ($p < 0.01$). We
want to note that both the proposed methods employed the simplest and most non-robust similarity metric,
SAD, while \emph{SPM\_D}, \emph{IRTK}, \emph{ART} and \emph{SyN} use advanced metrics (see \cite{klein2009evaluation}).
The choice of similarity metric was intentionally chosen for these experiments to demontrate the potential of
the proposed method without better similarity metrics or an advanced optimizer (i.e. a Levenberg-–Marquardt
optimizer as used in \emph{SPM\_D} \cite{ashburner2007fast}).

\subsection{Future directions}
The current \emph{RANCOR} framework can be seen as a basic method to be extended over time, under the
same open science credo, that allowed us to readily and quantitatively compare well-known open methods
using public databases. As the current framework cannot currently guarantee diffeomorphic deformations,
the next step is to enforce such constraints on the resulting deformation fields. Furthermore, to
enable inter-modality NLR, we will implement and test commonly used advanced similarity metrics,
such as normalized mutual-information, normalized cross-correlation, or more recently developed
methods, such as the $L2-norm$ of the MIND descriptor \cite{heinrich2012mind}. Since command-line
tools, such as the compared open NLR methods are needed for large-scale data analysis,
\emph{RANCOR} will be definitely included into such a package and, as a matter of course,
be made available to the community.

Additionally, we plan to extend our evaluation to include the RANCOR algorithm under an $L2$-norm regularization as a substitute for total variation as used in Sun et al\cite{sun2013efficient}. Such evaluation would allow for rigorous comparisons to be made by isolating the regularization mechanism.

\section{Conclusions}
We proposed a novel GPGPU-accelerated registration algorithm that optimizes any pointwise similarity
metric and total variation regularization within a Gauss-Newton optimization framework. This algorithm
was then evaluated against the four highest ranking non-linear registration algorithms according to
\cite{klein2009evaluation} on an open image database. We intend to provide our implementation back
to the community in an open manner.

\section*{Acknowledgments}

\appendix

\section{Dual Optimization Analysis}
\label{sec:proof01}

Gven the conjugate representation of the absolute function:
\begin{align}
\abs{v} \, = \, \max_{w} \, w \cdot v \, , \quad \text{s.t.} \; \abs{w} \, \leq
\, 1 \, ,
\end{align}
we can rewrite the first $L_1$-norm term of \eqref{eq:las} as
follows:
\begin{align} 
\int_{\Omega} \abs{P_0 + \nabla P \cdot h} dx   \,
= \, \max_{\abs{w(x)} \leq 1} \int_{\Omega} w (P_0 + \nabla P \cdot h) dx   
\label{eq:04}  \, . 
\end{align}
Additionally, given $R(\tilde{u}+h)$ in terms of
\eqref{eq:lreg}, we have
\begin{align} \label{eq:05}
& {\alpha} \sum_{i=1}^3 \int_{\Omega} \abs{\nabla (\tilde{u}_i + h_i)} dx
 = \max_{q} \, \sum_{i=1}^3 \int \diver q_i (\tilde{u}_i + h_i) dx  - 
 R^{\ast}(q)\, ,
\end{align}   
where each dual variable $q_i(x)$, 
$i=1,2,3$, has characteristic function
 of the constraint $\abs{q_i(x)} \leq \alpha$, $i=1,2,3$:
\bq \label{eq:dual-r1}
R^{\ast}(q) \, = \, \chi_{\abs{q_{1,2,3}(x)} \leq \alpha}(q) \, .
\eq

Considering \eqref{eq:04} and \eqref{eq:05}, one can see that
the convex minimization problem \eqref{eq:la} is equivalent to the minimax problem:
\begin{align}
 \min_h \max_{\abs{w(x)}\leq 1, q} \; & \int w (P_0 + \nabla P
\cdot h) dx + \, \sum_{i=1}^3 \int \diver q_i (\tilde{u}_i + h_i) dx
\, - \, R^{\ast}(q) \, 
\end{align}
that is
\begin{align}        
\hspace{-0.4cm}\min_h \max_{\abs{w(x)}\leq 1, q} \, & \int (w P_0 + \sum_{i=1}^3
\tilde{u}_i \diver q_i) dx + \, \sum_{i=1}^3 \int h_i(w \cdot
\partial_i P + \diver q_i) dx \, - \, R^{\ast}(q)  \label{eq:primal-dual}
\end{align}   
which is called the \emph{primal-dual formulation} in this paper.

After variation by the free variable $h_i(x)$, $i=1,2,3$, the minimization of
the \emph{primal-dual formulation} \eqref{eq:primal-dual} over $h_i(x)$,
$i=1,2,3$, results in the linear equalities' constraints
\begin{align}   
(w \cdot \partial_i P  +  \diver q_i)(x)  =  0 \, , \quad i = 1,2,3 \, ,
\end{align}
and the maximization problem
\begin{align} 
\nonumber \hspace{-0.4cm}\max_{\abs{w(x)}\leq 1, q} \, E(w,q) := & \int (w P_0 
+ \sum_{i=1}^3 \tilde{u}_i \diver q_i) dx \, - \, R^{\ast}(q) \,  
\end{align}
thereby proving Prop.~\ref{prop:01}.



\bibliographystyle{IEEEtran}
\bibliography{refs}
\end{document}